# On Monotonic Hamming Sphere Landscapes

DOMINIC A. WILSON

University of Findlay, Findlay, Ohio, USA, dominic.wilson@findlay.edu

Monotonic Hamming Sphere Landscapes (MHSL) are introduced: binary landscapes whose fitness is nonincreasing with Hamming distance from a unique global optimum, while points on the same Hamming sphere may be ordered arbitrarily. The class includes OneMax, every nondecreasing unitation function with a unique optimum, and positive linear functions with sufficiently balanced weights. The one-to-one MHSL, called Hamming sphere encodings (HSE), are counted in total and up to hypercube symmetry and shown to be a vanishing fraction of all landscapes. For the elitist (1+1) EA, every MHSL in which no two spheres share a fitness value has the optimization-time distribution of OneMax, which with known results makes these landscapes the easiest for mutation rates below one half. The same equivalence and optimality are proved for a two-member genetic algorithm with tournament selection and no crossover, whose hitting time on these landscapes reduces to a chain with quadratically many states. All 40,320 one-to-one three-bit landscapes are then evaluated for the (1+1) EA and the two-member genetic algorithm under eight mutation rates, three crossover operators and two selection schemes. Below one half, the easiest symmetry classes are all and only the HSE; above one half they are the hardest. Differences among HSE are small and stem from selection and from the hypercube symmetries respected by mutation and crossover. Targeted five-bit experiments likewise find their fastest landscapes within the HSE class.

CCS Concepts: • Theory of computation → Theory of randomized search heuristics; • Computing methodologies → Genetic algorithms; • Mathematics of computing → Markov processes.

**Additional Keywords and Phrases:** fitness landscapes, problem difficulty, Hamming distance, monotone functions, OneMax, Markov chains, genetic algorithms, symmetry, representations

## 1 INTRODUCTION

Fitness landscapes provide a foundational framework for understanding the difficulty of optimization problems encountered by evolutionary algorithms (EAs) (Malan & Engelbrecht, 2013; Malan, 2021). A central question in evolutionary computation is which landscapes are inherently easy or hard to optimize, and why. Decades of work on ruggedness, deception, neutrality and epistasis have produced many measures of difficulty, and a line of theory has identified the easiest and hardest functions for simple EAs (Doerr, Johannsen, & Winzen, 2012; Sudholt, 2013; He, Chen, & Yao, 2015). What is still missing is a clear picture of how the *arrangement* of fitness values on the hypercube, for a fixed set of values, governs difficulty for population-based algorithms with crossover.

In this paper we study **Monotonic Hamming Sphere Landscapes (MHSL)**, the binary landscapes in which fitness is ordered by Hamming distance from a unique global optimum x*. Points at the same distance from x* form a *sphere* (a Hamming sphere about x*). In an MHSL, every point on a sphere is at least as fit as every point on any sphere farther from the optimum; the definition does not constrain the ordering of points within a sphere. OneMax provides a familiar example: all points on a sphere have the same fitness, whereas MHSL allow arbitrary ordering within a sphere while preserving the fitness ordering between spheres.

When all fitness values are distinct, an MHSL also defines an encoding of phenotype values by the bit strings. We call this subclass Hamming sphere encodings (HSE). Standard binary and Gray coding are encodings in the same sense, though they are not generally HSE (Whitley, 1999; Chakraborty & Janikow, 2003a). Most studies of representation fix a phenotype-level function and compare the landscapes that selected encodings induce. We take the complementary view: fixing the hypercube, the algorithm and the set of fitness values, we ask which assignments of values to bit strings induce the easiest landscapes. For the three-bit problem studied exhaustively here, the answer is the HSE class; the larger instances test how far that answer extends.

The motivation for the HSE class comes from an exhaustive evaluation of all 40,320 three-bit landscapes. Among these landscapes, those that a two-member genetic algorithm solves fastest are neither standard binary nor Gray coded: they are all and only the HSE. This paper reports and explains that observation, separates what is already known from what is new, and tests how far it extends. Our contributions are:

1. We distinguish MHSL, sphere-separated MHSL and Hamming sphere encodings (HSE, the one-to-one case), and relate them to monotone and unitation functions (Section 2).
2. We enumerate HSE, give the orbit-counting argument behind the count of distinct neighborhoods, and show which symmetry group governs each algorithm. The same argument re-derives the structure of the 3-bit tables of Chakraborty and Janikow (2003b) (Section 4).
3. We show that for the (1+1) EA, every MHSL in which no two spheres share a fitness value has the same hitting-time distribution as OneMax. Combined with known results, this makes them the easiest landscapes for mutation rates below 0.5. We also explain why the order inverts above 0.5 (Section 5).
4. We prove that for a two-member GA with binary tournament selection and no crossover, every sphere-separated MHSL again has OneMax's distance chain and is an easiest landscape for $p < 0.5$, and that with uniform crossover all HSE share one chain on $O(n^3)$ states (Section 5).
5. We analyze a two-member genetic algorithm with Markov chains on all 40,320 three-bit landscapes. Under fitness-proportional and tournament selection, with and without crossover, the easiest classes in every tested setting below 0.5 are all and only the HSE, and above 0.5 the hardest are (Section 7).
6. We explain the differences within the HSE class through selection and operator symmetry: tournament selection removes the differences due to numerical fitness values (Propositions 3 and 4), uniform crossover preserves the full symmetry of the cube, and one-point crossover introduces positional distinctions that split the 6 classes into 18. Under mutation alone, the differences track how the values on adjacent spheres are aligned, with a sign that flips between the spheres nearest the optimum and the rest (Section 7.2).

## 2 BACKGROUND AND DEFINITIONS

### 2.1 Landscapes

Let n be the number of binary variables, so the search space is $\{0,1\}^n$ with $2^n$ points, and let $f : \{0,1\}^n \to \mathbb{R}$ be a fitness function to be maximized (without loss of generality). The landscape is the pair $(\{0,1\}^n, f)$. It can be drawn as the n-dimensional hypercube whose vertices are bit strings and whose edges join strings at Hamming distance 1; f(x) labels the vertex x. If the $2^n$ fitness values are distinct, they can be assigned in $(2^n)!$ ways, so there are 8! = 40,320 landscapes on three bits. Throughout, the fitness values are the integers 0, 1, …, $2^n - 1$, and the optimum has value $2^n - 1$.

## 2.2 Hamming spheres and monotonicity

Fix a unique global optimum x* and let d(x) = H(x, x*) be the Hamming distance to it. The *sphere* at distance k is $S_k$ = {x : d(x) = k}, which contains C(n,k) points, for $0 \le k \le n$.

**Definition 1 (Monotonic Hamming Sphere Landscape).** A fitness function f is a Monotonic Hamming Sphere Landscape (MHSL) with optimum x* if, for all x and y,

$$d(x) < d(y) \;\Rightarrow\; f(x) \ge f(y).$$

We use **weak MHSL** when emphasizing that inter-sphere ties are permitted.

**Definition 2 (sphere-separated MHSL).** An MHSL f is sphere-separated if fitness ties are allowed inside a sphere but not across spheres, that is:

$$d(x) < d(y) \;\Rightarrow\; f(x) > f(y).$$

**Definition 3 (Hamming sphere encoding, HSE).** An HSE is an injective (one-to-one) MHSL whose fitness values are {0, …, 2^n − 1}; injectivity makes it sphere-separated.

## 2.3 Relation to known classes

Unitation functions (fitness depends only on the number of ones) are *flat*, meaning that they carry a single fitness value, on every sphere about $1^n$. With x* = $1^n$, the MHSL with flat spheres are exactly the unitation functions that are nondecreasing in the number of ones and have a unique maximum. These functions form a proper subclass of MHSL. A nondecreasing unitation function whose maximum is not unique, such as f(x) = min(|x|, n − 1), is not an MHSL.

OneMax is an example of a unitation function that is an MHSL. It is sphere-separated, with x* = $1^n$ and f(x) = n − d(x), and all of its spheres are flat. Monotone step and threshold functions of Hamming distance, when defined to have a unique optimum, are weak MHSL. Needle in a Haystack, with f(x*) = 1 and f(x) = 0 otherwise, is a weak MHSL in which every sphere except $S_0$ is tied.

With x* = $1^n$, every sphere-separated MHSL is a strictly monotone pseudo-Boolean function: if y is obtained from x by flipping a 0 to a 1, then d(y) < d(x) and therefore f(y) > f(x). The converse fails, because monotonicity compares only strings whose 1-bits are nested, whereas an MHSL compares every pair of strings at different distances from the optimum. For example, BinVal is strictly monotone, yet it gives 1000 the value 8 and 0111 the value 7, although 0111 is at distance 1 from the optimum 1111 and 1000 is at distance 3. MHSL also impose more than unimodality. A unimodal landscape need only give every non-optimal point an improving neighbor and may still contain an improving path of exponential length (Horn, Goldberg, & Deb, 1994); in a sphere-separated MHSL every move to a closer sphere improves fitness and every move to a farther sphere worsens it, which aligns fitness with distance globally rather than merely excluding non-global local optima. The distinction matters for the comparisons in this paper. For the phenotype values 0, …, $2^n - 1$, standard binary, reflected Gray coding and every HSE all induce landscapes with a single local optimum under single-bit flips: in an HSE every non-optimal point has a fitter neighbor on the next closer sphere, under standard binary flipping the highest-order 0 bit increases the represented value, and under reflected Gray coding each non-maximal value has its successor at Hamming distance one. A count of local optima therefore cannot separate them, whereas their expected hitting times differ (Section 7.6).

MHSL also intersect nontrivially with the class of positive linear functions $f(x) = \sum_{i=1}^{n} w_i x_i$, where $w_i > 0$. Such a function is a weak MHSL precisely when, for every $k < n$, the sum of the $k + 1$ smallest weights is at least the sum of the $k$ largest weights. It is sphere-separated when each of these inequalities is strict. Equivalently, every solution with $k + 1$ one-bits must be at least as fit as every solution with $k$ one-bits, and strictly fitter in the sphere-separated case. Because the weights are positive, the optimum is $1^n$. Equal weights give OneMax.

A simple sufficient condition for sphere separation is that every weight be less than $n/(n-1)$ times the smallest, $w_{max} < n \cdot w_{min}/(n-1)$: then the $k+1$ smallest weights sum to at least $(k+1)w_{min} > k \cdot w_{max}$, which bounds the sum of the k largest, for every $k < n$. The weights 1, 1 and 1.5 meet the exact condition, though not the sufficient one, whereas the exponentially increasing weights of BinVal (1, 2 and 4 for $n = 3$) fail it at $k = 1$.

Several standard benchmarks are *not* MHSL. LeadingOnes, which is only weakly monotone, assigns 0111 lower fitness than 1000 although 0111 is closer to the optimum 1111. Jump, Cliff and Trap functions violate sphere order by construction, and Trap is its maximal violation (Section 7.3). Lengler and Riedi's Dynamic BinVal is a time-varying linear function and likewise falls outside the class. Monotone functions are well studied (Doerr, Jansen, Sudholt, Winzen, & Zarges, 2013; Lengler, 2020); what is specific to MHSL is the free order inside spheres, which is the object of Sections 4 and 7.

## 3 RELATED WORK

**Landscape structure and difficulty.** The landscape metaphor goes back to Wright (1932). In evolutionary computation, tunable landscapes such as NK models (Kauffman, 1993) and deceptive functions (Goldberg, 1987) showed how epistasis and misleading gradients make search hard, and local optima networks map the basins that cause it (Ochoa, Tomassini, Vérel, & Darabos, 2008). Jones and Forrest (1995) proposed fitness-distance correlation (FDC), the correlation between fitness and distance to the optimum, as a predictor of GA difficulty. Malan and Engelbrecht (2013) and Malan (2021) survey the many landscape measures proposed since. MHSL are the landscapes in which fitness order and sphere order agree perfectly, the agreement that FDC measures on average; Section 7.4 uses a rank version of it to measure how far a landscape is from the class.

**Easiest and hardest functions.** For the (1+1) EA, OneMax is the easiest function with a unique optimum (Doerr et al., 2012; Sudholt, 2013). He, Chen and Yao (2015) characterize the easiest and hardest functions in a class with common optima for an elitist (1+1) EA: a function is easiest when fitness is ordered consistently with expected hitting time, and hardest when the order is reversed, which yields deceptive (trap-like) functions. They also note that the easiest function depends on the algorithm. Kaufmann, Larcher, Lengler and Zou (2025) show that OneMax is easiest for reducing distance to the optimum but not for finding fitness-improving steps. Unimodality alone does not make a function easy: some unimodal functions take exponential time (Droste, Jansen, & Wegener, 2002). Non-elitist populations are covered by the negative-drift method of Lehre (2010), simplified by Doerr (2021), which establishes exponential lower bounds under conditions relating mutation rate, selection and reproductive rate. Section 5 places MHSL inside this theory and extends the easiest-function result to a two-member GA with tournament selection; Section 7 examines settings not covered by that result, including fitness-proportional selection and crossover.

**Markov models and representations.** Finite Markov chain models of the simple GA (Nix & Vose, 1992; Vose, 1999) give analytically determined expected hitting times for small problems. Chakraborty and Janikow (2003a, 2003b) used them to compare binary and Gray coding over all 40,320 three-bit functions, with a two-member population, and found little overall difference, as the no-free-lunch theorem predicts for comparisons over all functions (Wolpert & Macready, 1997); Whitley (1999) and Rowe, Whitley, Barbulescu and Watson (2004) identify the function classes on which Gray coding does win. Exhaustive or near-exhaustive Markov-chain studies of small problems go back to Goldberg and Segrest (1987), De Jong, Spears and Gordon (1995) and Spears and De Jong (1997), and Rana and Whitley (1997) counted local optima under binary and Gray neighborhoods over all small functions. Our method follows theirs. The difference is the question: instead of comparing two fixed encodings across all functions, we ask which encodings of a fixed set of fitness values are best. Nontrivial average-case representation comparisons therefore require restricting attention to structured subsets of functions, because all representations are equivalent when averaged uniformly over all functions (Radcliffe & Surry, 1995). Whitley and Rana (1997) condition on the number of local optima in a defining neighborhood and show that Gray coding then induces fewer local optima than standard binary on average; our exhaustive study instead identifies agreement between fitness order and Hamming distance as the structure shared by the easiest encodings.

**Representation equivalence.** Rowe, Whitley, Barbulescu and Watson (2004) call two encodings bit equivalent when they differ by an automorphism of the hypercube, and show that bit equivalence preserves neighborhood structure and the magnitudes of the Walsh coefficients. Section 4 refines this for stochastic search: the encodings guaranteed to share the expected hitting time under the uniform initial distribution are the orbits of the subgroup of automorphisms that the whole transition kernel respects, so changing the crossover operator changes which encodings are equivalent on the same hypercube.

**Symmetry.** Rowe, Vose and Wright (2002, 2004, 2010) describe when crossover and mutation commute with a group acting on the search space, and which operators make a GA invariant under a change of representation. Section 4.2 applies this to expected hitting times and uses it to explain the structure of the three-bit results.

## 4 ENUMERATION AND SYMMETRY OF HAMMING SPHERE ENCODINGS

### 4.1 Counting

**Proposition 1.** The number of HSE on n bits is

$$E_n = 2^n \cdot \prod_{k=0..n} C(n,k)!,$$

and the number of HSE that are distinct up to the symmetries of the hypercube is $\prod_k C(n,k)! \,/\, n!$.

**Proof.** With x* fixed, the fitness values assigned to $S_k$ are forced (the C(n,k) values just below those of $S_{k-1}$), and they can be arranged inside $S_k$ in C(n,k)! ways, independently for each sphere. The optimum can be placed at any of the $2^n$ vertices. The automorphism group $B_n$ of the hypercube (bit permutations composed with bitwise XOR masks) has order $2^n \cdot n!$ and maps HSE to HSE. Because the fitness values are distinct, only the identity fixes a landscape, so the action is free and every orbit has $2^n \cdot n!$ elements. ■

For n = 3 there are 288 HSE, forming the 6 classes listed in Table 1, out of 40,320 landscapes (0.71%). For n = 4 there are 6,635,520 HSE in 17,280 classes, a fraction of $3.2 \times 10^{-7}$ of all 16! landscapes; for n = 5 the fraction is $2.3 \times 10^{-17}$. The number of HSE grows doubly exponentially (Appendix B), but so does the number of all landscapes, and the more informative statement is that HSE form a vanishing fraction of them.

Table 1 lists the six HSE classes for n = 3 as encodings: the row for decimal label r gives the bit string that receives value r. Standard binary and binary reflected Gray coding (BRGC) are shown for comparison. All six place the values 1, 2 and 3 on the strings with a single one and the values 4, 5 and 6 on the strings with two ones; they differ only in the order inside the middle spheres.

Table 1: Standard binary code, binary reflected Gray code and the 6 classes of Hamming sphere encodings for n = 3. Each of the 6 classes contains 48 of the 288 HSE.

| Decimal label | Standard binary | Gray (BRGC) | HSE1 | HSE2 | HSE3 | HSE4 | HSE5 | HSE6 |
|---|---|---|---|---|---|---|---|---|
| 0 | 000 | 000 | 000 | 000 | 000 | 000 | 000 | 000 |
| 1 | 001 | 001 | 001 | 001 | 001 | 001 | 001 | 001 |
| 2 | 010 | 011 | 010 | 010 | 010 | 010 | 010 | 010 |
| 3 | 011 | 010 | 100 | 100 | 100 | 100 | 100 | 100 |
| 4 | 100 | 110 | 011 | 101 | 011 | 110 | 101 | 110 |
| 5 | 101 | 111 | 101 | 011 | 110 | 011 | 110 | 101 |
| 6 | 110 | 101 | 110 | 110 | 101 | 101 | 011 | 011 |
| 7 | 111 | 100 | 111 | 111 | 111 | 111 | 111 | 111 |

### 4.2 Which symmetries does an algorithm respect?

The count above uses the full group $B_n$, but an algorithm need not treat all of $B_n$ alike. Let G be a group of permutations of $\{0,1\}^n$ acting on landscapes by $(g \cdot f)(x) = f(g^{-1}x)$.

**Lemma 1.** Suppose the mutation and crossover kernels commute with every $g \in G$ (for example $M(gx, gy) = M(x, y)$), selection depends only on fitness values, and the initial population is drawn uniformly. Then the expected hitting time of the optimum is the same for f and $g \cdot f$, for every $g \in G$.

*Proof sketch.* Applying g to every individual maps populations for f to populations for $g \cdot f$, carries the transition probabilities of one chain onto those of the other, maps the absorbing set (populations containing $x^*$) onto the absorbing set for $g \cdot f$ (populations containing $gx^*$), and preserves the uniform initial distribution. The fundamental matrices are therefore permutations of each other. This is the hitting-time form of the invariance results of Rowe et al. (2002, 2010). ■

Two groups matter here. Bitwise mutation and uniform crossover commute with all of $B_n$ (order $2^n \cdot n!$, 48 for n = 3). One-point crossover depends on which bits are adjacent, so among bit permutations it commutes only with the identity and the reversal of bit order; together with XOR masks this gives a subgroup $H_n$ of order $2^{n+1}$ (16 for n = 3). On three bits the 40,320 landscapes form 840 orbits under $B_3$ and 2,520 under $H_3$, and the 6 HSE classes split into 18 $H_3$-classes. Selection adds *value* symmetries: rank-based selection, binary tournaments and the elitist (1+1) EA are invariant under any order-preserving change of fitness values, while fitness-proportional selection is not. Section 7.2 confirms each of these predictions numerically.

### 4.3 The three-bit tables of Chakraborty and Janikow

The same reasoning explains the tables of Chakraborty and Janikow (2003b). Their Table 3 reports identical numbers of functions with 1, 2, 3 and 4 local minima under binary and Gray neighborhoods (8,640, 24,480, 5,760 and 1,440). This needs no experiment. A bijective encoding only permutes the set of all 8! functions, and Gray coding carries the binary cube onto the Gray cube, so the distribution of minima counts must be the same.

Their Table 4 groups the functions into 26 classes by the number of local minima in the integer (ring), Gray and binary neighborhoods. We recomputed all 26 class sizes (Appendix C, Table 10), and every one is divisible by 16. The reason is a group of order 16 acting freely. On the position side, the permutations of $\{0, \ldots, 7\}$ that are automorphisms of the ring, the binary cube and the Gray cube at the same time form the Klein four-group $\{x, x \oplus 3, x \oplus 4, x \oplus 7\}$. On the value side, exchanging the values 1 and 2, or 7 and 8, never changes which points are local minima: 1 is always a minimum, 2 is a minimum if and only if it is not adjacent to 1, and 7 and 8 can never be minima. Their win counts (Gray versus binary) are divisible only by 4. Their GA uses fitness-proportional selection, which depends on the actual values, so only the position symmetries survive.

## 5 ONEMAX EQUIVALENCE AND REDUCTIONS FOR THE (1+1) EA AND THE TOURNAMENT GA

### 5.1 The (1+1) EA

Consider the elitist (1+1) EA with standard bit mutation at rate $p \in (0, 1)$, which replaces the current point x by its offspring y whenever $f(y) \geq f(x)$.

**Proposition 2.** On any sphere-separated MHSL with optimum $x^*$, the distance process $D_t = d(X_t)$ of the (1+1) EA is a Markov chain with the same transition probabilities as on OneMax. In particular, the optimization time has the same distribution as on OneMax from every starting point at the same distance.

**Proof.** Let $d(x) = d$. Bitwise mutation is invariant under the symmetries of the cube that fix $x^*$, so the probability that the offspring lies at distance $d'$ depends only on d and $d'$, and it is the same for every landscape. If $d' < d$, the offspring is strictly better and is accepted, as on OneMax. If $d' > d$, it is strictly worse and is rejected, as on OneMax. If $d' = d$, it may or may not be accepted, but the distance does not change either way. Hence the law of $D_{t+1}$ given the past depends only on $D_t$, with the same kernel as for OneMax. ■

**Corollary 1.** For $p < 0.5$, every sphere-separated MHSL is an easiest function for the (1+1) EA among functions with the same optimum.

**Proof.** By Proposition 2, the expected hitting time G(x) from x equals OneMax's value g(d(x)) at distance d(x). By Lemma 2 below, g is strictly increasing for $p < 0.5$. So $G(x) < G(y)$ implies $d(x) < d(y)$, which implies $f(x) > f(y)$, and the easiest-function criterion of He et al. (2015, Theorem 1) applies. ■

Easiest in Corollary 1 means smallest expected optimization time, the sense used by Doerr et al. (2012) and He et al. (2015). Kaufmann et al. (2025) show that in a different sense OneMax is not easiest: for the (1, λ) EA with a self-adjusting population size, what matters is the probability that an offspring improves fitness, and other functions make improvements easier to find. Their observation does not conflict with Corollary 1: on a sphere-separated MHSL, every move to a closer sphere improves fitness and every move to a farther sphere worsens it, as on OneMax, although moves within a sphere may be ordered differently; the two notions of easiness simply measure different things.

**Lemma 2.** Let g(d) be the expected optimization time of the (1+1) EA with mutation rate p on OneMax from a point at distance d from the optimum. If $p < 0.5$, then $g(0) < g(1) < \ldots < g(n)$. If $p = 0.5$, g is constant on $\{1, \ldots, n\}$. For $p > 0.5$ the monotonicity can fail (Table 3 and the remarks below).

**Proof.** Fix $0 \leq d < n$, let x be at distance d, and let y be obtained from x by flipping one correct bit i, so that y is at distance $d + 1$. Write W for the d bits that are wrong in x and R for the $n - d - 1$ bits other than i that are right in x. Let S be the mutation mask applied to x and $S'$ the mask applied to y. We couple them as follows: on every bit other than i, $S' = S$; at bit i, at most one of the two masks flips i, S alone with probability p, $S'$ alone with probability p, and neither with probability $1 - 2p$. This coupling is valid if and only if $p \leq 0.5$, and it gives each mask its correct marginal distribution. Let $a = |S \cap W|$ and $b = |S \cap R|$. Before selection, the mutated offspring of x and y lie at distances $d_{prop}(x) = d - a + b + [i \in S]$ and $d_{prop}(y) = d + 1 - a + b - [i \in S']$. Because $[i \in S] + [i \in S'] \leq 1$ under the coupling, we have $d_{prop}(x) \leq d_{prop}(y)$ pathwise, which shows that the unconstrained mutation distance kernel is stochastically monotone for $p \leq 0.5$. Under elitist selection on OneMax, each offspring is accepted if and only if its distance does not exceed its parent's, so the new distances are $D_x = \min(d, d_{prop}(x))$ and $D_y = \min(d + 1, d_{prop}(y))$. Since $d \leq d + 1$ and $d_{prop}(x) \leq d_{prop}(y)$, it follows immediately that $D_x \leq D_y$ in every case. Hence the one-step kernel of the elitist distance chain is also stochastically monotone: from a larger distance the next distance is stochastically larger. For a stochastically monotone chain, runs from d and from d + 1 can be coupled so that the second is never closer to 0 than the first at any time (Lindvall, 1992, Chapter IV), so the hitting time of 0 from d + 1 is at least that from d pathwise and $g(d) \leq g(d + 1)$. For strictness with $p < 0.5$, the event that S flips exactly the bits of W and neither mask flips i has positive probability; on it the run from d is absorbed at the first step while the run from d + 1 is not. On this event the first hitting times differ strictly, while under the coupling the hitting time from d + 1 is never smaller; hence the expected hitting times differ strictly, so $g(d) < g(d + 1)$. For $p = 0.5$ the offspring is uniform regardless of the parent, so g is constant away from 0. For $p > 0.5$ the coupling no longer exists. ■

Numerical checks for $n \leq 14$ were consistent with the proved monotonicity for $p \leq 0.5$, while counterexamples appeared at $p = 0.51$ (specifically at $n = 3$).

Three remarks follow. First, inter-sphere ties matter: on a weak MHSL with ties across spheres, an offspring that is farther away but equally fit is accepted, so the (1+1) EA can drift away from the optimum and the equivalence with OneMax fails. This is the precise sense in which neutrality between spheres, not the free order inside them, can make an MHSL harder. Second, for $p > 0.5$ OneMax's expected hitting time decreases with distance (for $n = 5$ and $p = 0.65$ it falls from 102.5 steps at distance 1 to 87.5 at distance 5), because a mutation is then likely to land near the complement of the parent. The criterion of He et al. then reverses, and Section 7.5 shows that HSE become the hardest landscapes. Third, in our three-bit data every landscape that is not an HSE is strictly slower than the HSE for $p < 0.5$, which suggests the converse:

**Conjecture 1.** For $p < 0.5$, an injective landscape with a unique optimum attains OneMax's expected hitting time from every starting point only if it is a Hamming sphere encoding.

### 5.2 Tournament selection without crossover

The argument extends to the GA of Section 6 whenever selection only compares fitness values. Write the population as (x, y) and let $D_t$ be the pair (d(x), d(y)) sorted in increasing order. A binary tournament draws two members with replacement and keeps the fitter, so the fitter member of the population is selected with probability 3/4 and, when the two are equally fit, each with probability 1/2.

**Proposition 3.** Consider the two-member GA with binary tournament selection and bitwise mutation at rate $p \in (0, 1)$. Without crossover, on every sphere-separated MHSL the process $D_t$ (the sorted distance pair of the population) is a Markov chain with the same transition probabilities as on OneMax; it has $(n + 1)(n + 2)/2$ states. The expected hitting time is the same for every landscape in the stated class from initial populations having the same reduced state, and therefore also when averaged over uniformly distributed initial populations.

**Proof.** On a sphere-separated MHSL, $d(x) < d(y)$ implies $f(x) > f(y)$, so each parent draw selects the closer member with probability 3/4 and the farther with probability 1/4; when $d(x) = d(y)$ the draw may favor either member, but both are at the same distance. Mutation is invariant under the symmetries of the cube that fix $x^*$, so the distance of a mutant depends only on the distance of its parent, and the two children are mutated independently. Hence the law of $D_{t+1}$ given the past depends only on $D_t$, through n and p alone, and the run stops exactly when $\min D_t = 0$. Finally, under a uniform initial population the distribution of $D_0$ is the same for every landscape with a unique optimum. ■

### 5.3 Uniform crossover

**Proposition 4.** Consider the two-member GA with binary tournament selection and uniform crossover at any rate $p_c > 0$, with bitwise mutation at rate $p \in (0, 1)$. On every HSE, let $C_t$ be the equivalence class of $(c_{11}, c_{10}, c_{01})$ under exchange of $c_{10}$ and $c_{01}$, represented canonically by taking $c_{10} \leq c_{01}$, where $c_{11}$ is the number of positions at which both members differ from $x^*$, $c_{10}$ the number at which only the first does and $c_{01}$ the number at which only the second does. Then $C_t$ is a Markov chain with the same transition probabilities for every HSE; it has at most $(n + 1)(n + 2)(n + 3)/6$ states. The expected hitting time is the same for every HSE from initial populations having the same reduced configuration state, and therefore also when averaged over uniformly distributed initial populations.

**Proof.** The bit permutations that fix x* act transitively on the ordered pairs with a given configuration, and uniform crossover and mutation commute with them, so the configuration equivalence class of the two children depends only on the ordered configuration of the selected parents, and is further invariant under exchanging the two parents because uniform crossover and bitwise mutation treat both parents symmetrically. On an HSE there are no fitness ties between distinct strings, so the fitter member of the current population is selected with probability 3/4 and the less fit with probability 1/4; homogeneous mating of the fitter member, heterogeneous mating of the two members, and homogeneous mating of the less fit member therefore occur with probabilities 9/16, 6/16 and 1/16. The two ordered mixed parent pairs each have probability (3/4)(1/4) = 3/16, giving total heterogeneous-mating probability 6/16. With probability 1 − p_c, crossover is not applied and the selected parents are copied directly before mutation; this branch also depends only on their canonical configuration and is invariant under exchanging them.

In the canonical representation $C_t = (c_{11}, c_{10}, c_{01})$ with $c_{10} \leq c_{01}$, the member with distance $c_{11} + c_{10}$ is strictly closer and hence strictly fitter whenever $c_{10} < c_{01}$, so the three mating types have canonical parent configurations $(c_{11} + c_{10}, 0, 0)$, $C_t$, and $(c_{11} + c_{01}, 0, 0)$. When $c_{10} = c_{01}$, both members lie at the same distance $d = c_{11} + c_{10}$; homogeneous mating produces the same canonical parent configuration (d, 0, 0) from either equally distant member, while heterogeneous mating has canonical parent configuration $C_t$ and uniform crossover is invariant under exchanging its parents. Thus, although different HSE may reverse the fitness ordering of two members on the same sphere, that reversal only exchanges the labels "fitter" and "less fit" and leaves the canonical offspring-configuration distribution unchanged. Hence the law of $C_{t+1}$ depends only on $C_t$ in all cases. Without identifying configurations related by exchanging the two members, the number of nonnegative triples satisfying $c_{11} + c_{10} + c_{01} \leq n$ is $\binom{n+3}{3} = \frac{(n+1)(n+2)(n+3)}{6}$; canonical ordering can only reduce this number. Finally, under a uniform initial population the distribution of $C_0$ is the same for every HSE. ■

Proposition 3 explains why all sphere-separated MHSL, including all HSE, have the same reduced-chain hitting time under tournament selection without crossover. It also makes the tournament GA computable for any n: the full chain has $4^n$ states, the reduced chain of Proposition 3 has $O(n^2)$ and that of Proposition 4 has $O(n^3)$. At p = 1/n the chain of Proposition 3 needs 15.0 generations for n = 5, 181 for n = 10, $1.6 \times 10^4$ for n = 20 and $1.3 \times 10^8$ for n = 40. The computed values grow rapidly with n, consistent with the exponential behavior known for related non-elitist population models (Lehre, 2010; Doerr, 2021).

Proposition 4 does not extend to ties inside spheres: with a tie, the mixed parent pair arises with probability 1/2 instead of 3/8, which changes the chain. This is the one setting in the paper where within-sphere ties affect the comparison with OneMax. A likely mechanism is that crossover can combine correct bits from two members of the same sphere, while ties allow such pairs to be selected more frequently. On OneMax, where every sphere is one tie, the tournament GA with uniform crossover needs 26.85 generations at n = 5, p = 0.05 and $p_c = 0.8$, against 27.55 on every HSE, and with one-point crossover 26.93 against 27.59; without crossover the two coincide, as Proposition 3 says.

### 5.4 Optimality of sphere-separated MHSL

**Lemma 3.** Let $h(d_1, d_2)$ be the expected hitting time from $(d_1, d_2)$, with $h(d_1, d_2) = 0$ whenever $\min(d_1, d_2) = 0$. The state space consists of sorted pairs $(d_1, d_2)$ with $0 \leq d_1 \leq d_2 \leq n$. For $p \leq 0.5$, h is nondecreasing in each coordinate on this state space. For every nonabsorbing state $d = (d_1, d_2)$, h satisfies the first-step equation $h(d) = 1 + E_{MHSL}[h(D_{t+1}) \mid D_t = d]$.

**Proof.** Let $(a_1, a_2) \leq (b_1, b_2)$ coordinatewise within the state space and run two copies of the chain from these states with coupled randomness. Each parent draw uses the same uniform variable in both copies to decide between the closer and the farther member, so each parent of the second copy is at least as far as the corresponding parent of the first. Each child is produced by the monotone coupling of the unconstrained mutation distance kernel, which exists for $p \leq 0.5$ because the kernel is stochastically monotone (as established in the proof of Lemma 2; Lindvall, 1992), so each child of the second copy is at least as far as the corresponding child of the first. Sorting preserves the coordinatewise order and keeps the state within bounds. The order is therefore kept at every generation, so the first copy reaches $\min D = 0$ no later than the second, and $h(a_1, a_2) \leq h(b_1, b_2)$. The same coupling compares selection strengths $q \leq q'$. Draw U uniformly on (0, 1) and let the q-chain select the closer member when $U \leq q$ and the q'-chain when $U \leq q'$; the parent chosen by the q'-chain is then never farther from the optimum than the parent chosen by the q-chain, and applying the monotone mutation coupling independently to each pair of corresponding offspring and then sorting preserves the coordinatewise order, so for $p \leq 0.5$ the expected time is nonincreasing in q. ■

**Theorem 1.** Let $p < 0.5$ and let g be any landscape on $\{0, 1\}^n$ with a unique global optimum $x^*$. For the two-member GA with binary tournament selection and no crossover, the expected number of generations to reach $x^*$ from any initial population (x, y) is at least $h(d(x), d(y))$, the value from a population at the same distances on any sphere-separated MHSL. Two consequences follow. Among landscapes with the same optimum, every sphere-separated MHSL minimizes the expected hitting time from every initial population. For a uniformly distributed initial population the sorted distance pair $D_0$ has the same law on every landscape with a unique optimum, so every sphere-separated MHSL also minimizes the average expected hitting time among all such landscapes, wherever their optima lie.

**Proof.** Let $(X_t, Y_t)$ be the population at generation t under g with filtration $F_t$, let $D_t$ be its sorted distance pair, and let $K(d, \cdot)$ be the law of the distance of a mutant whose parent is at distance d; for $p < 0.5$ it is stochastically increasing in d (Lemma 2). From a population at sorted distances $d_1 \leq d_2$ under g, if $d_1 < d_2$, a parent draw selects the closer member with probability $q \in \{1/4, 1/2, 3/4\}$, according to whether it is less fit, equally fit, or fitter. If $d_1 = d_2$, both potential parents have mutation-distance law $K(d_1, \cdot)$, so their fitness order is irrelevant to the child-distance distribution (equivalent to $q = 1/2$). On a sphere-separated MHSL, $q = 3/4$ whenever $d_1 < d_2$. Conditioned on $F_t$, each child's distance therefore has law $qK(d_1, \cdot) + (1 - q)K(d_2, \cdot)$, which is stochastically decreasing in q when $d_1 < d_2$ and independent of q when $d_1 = d_2$. Couple each child under g to the corresponding child of the MHSL chain started from $(d_1, d_2)$ through the one-dimensional stochastic order of their distance laws, the two couplings independent conditional on $(X_t, Y_t)$. Each child under g is then at least as far from the optimum as its MHSL counterpart, and sorting preserves the coordinatewise order, so $D_{t+1}$ under g stochastically dominates the next state of the chain of Proposition 3 in the coordinatewise order. Since h is coordinatewise nondecreasing on the whole state space, absorbing states included (Lemma 3), $E_g[h(D_{t+1}) \mid F_t] \geq h(D_t) - 1$ on the event $\{T > t\} = \{\min D_t > 0\}$, the right-hand side being the first-step equation of the MHSL chain. Thus $M_t = h(D_{t\wedge T}) + (t \wedge T)$ is an $F_t$-submartingale, where $T = \inf\{t \geq 0 : \min D_t = 0\}$. In every nonabsorbing generation the first child equals $x^*$ with probability at least $p^n$, since a parent at distance d produces $x^*$ with probability $p^d(1 - p)^{n-d} \geq p^n$ for $p < 1/2$; hence $P_g(T > t) \leq (1 - p^n)^t$ and $E_g[T] < \infty$. As h is bounded on the finite state space, $|M_t| \leq \max h + T$. Since $M_t$ is a submartingale, $E_g[M_t] \geq M_0$ for every t, and $M_t \to M_T = T$, so dominated convergence gives $E_g[T] = E_g[M_T] = \lim_{t\to\infty} E_g[M_t] \geq M_0 = h(D_0)$. ■

Theorem 1 is the GA counterpart of Corollary 1, and its proof is the criterion of He et al. (2015) lifted to the pair chain: Lemma 3 is the condition that fitness order and time order agree. By the last part of Lemma 3, on an MHSL the time falls as the closer member is selected more often: replacing 3/4 by q gives 83.7, 48.6, 28.7, 21.4 and 19.2 generations for q = 1/2, 0.6, 3/4, 0.9 and 1 at n = 5 and p = 0.05. Variation in effective selection strength is therefore one mechanism that separates the HSE under fitness-proportional selection, where q depends on the numerical fitness values (Section 7.2). For $p > 0.5$ the argument fails at Lemma 3: h is no longer monotone (at $n = 4$, 5 and 8 it already fails at $p = 0.55$). In comparisons with 200 random landscapes at $n = 4$ and 5 for $p = 0.65$, the HSE were slower in every case, consistent with the three-bit reversal reported in Section 7.5; we have no general proof. Conjecture 1 has the following GA counterpart, supported by the exhaustive three-bit results of Section 7.1 and the local searches of Section 7.6.

**Conjecture 2.** For $p < 0.5$, every global minimizer of expected hitting time among injective landscapes with a unique optimum is an HSE for the two-member GA under (a) tournament selection with uniform crossover and (b) fitness-proportional selection with no crossover or uniform crossover.

In case (a), all HSE share the same expected hitting time by Proposition 4, so the conjecture there says that the minimizers are exactly the HSE; in case (b) the HSE differ among themselves, and the conjecture says that every globally minimizing landscape belongs to the HSE class, not that every HSE is a minimizer. The exhaustive $n = 3$ results support this "every minimizer" formulation, whereas the larger-n searches of Section 7.6 found only HSE among the local optima of swap moves, which does not establish global optimality. The injectivity restriction in case (a) is necessary for the conjecture as stated because, in our tested crossover settings, OneMax is faster than every HSE.

Proposition 3 and Theorem 1 settle the tournament GA without crossover. The remainder of the paper concerns what they leave open, fitness-proportional selection and crossover, where the exact chains have to be evaluated.

## 6 EXPERIMENTAL METHOD

**Markov model.** We follow the Markov model of Nix and Vose (1992) as used by Chakraborty and Janikow (2003a, 2003b). The GA has a population of two individuals of n bits. In each generation, two parents are chosen by selection (with replacement), crossed with probability $p_c = 0.8$ to produce two children, and each child is mutated bitwise with probability p. Selection is either fitness-proportional, with the integers 0, …, $2^n - 1$ as fitness values, or a binary tournament (the fitter of two individuals drawn with replacement wins). Crossover is none ($p_c = 0$), one-point (cut point uniform over the $n - 1$ internal positions), or uniform. Every population that contains the optimum is made absorbing. The expected number of generations to absorption is obtained from the fundamental matrix $(I - Q)^{-1}$ and averaged uniformly over all $2^{2n}$ initial ordered populations, counting populations that already contain the optimum as zero. For the (1+1) EA we use the same construction on single points.

**Differences from Chakraborty and Janikow.** They minimize, use fitness values 1, …, 8 and keep one randomly chosen child per crossover. We maximize fitness values 0, …, 7 and keep both children. Absolute step counts therefore differ, but the symmetry arguments of Section 4 apply to both.

**Orbit reduction.** By Lemma 1 we evaluate one representative per orbit: 840 $B_3$-orbits for the (1+1) EA, mutation-only GA and uniform crossover, and 2,520 $H_3$-orbits for one-point crossover. Spearman correlations are computed over all 40,320 one-to-one landscapes by weighting each orbit by its size.

**Settings.** For $n = 3$ we evaluate a full grid of 56 settings: eight mutation rates $p \in \{0.02, 0.05, 0.1, 0.2, 1/3, 0.45, 0.55, 0.65\}$ for each of seven algorithm variants, namely the (1+1) EA and the GA with each of two selection schemes and three crossover options. The rates cover typical GA settings (0.02 to 0.1), the standard (1+1) EA rate $1/n = 1/3$, two rates that bracket 0.5, and one well above it. For $n = 4$ to 6, Section 7.6 reports four targeted tests: a comparison with other encodings, perturbations of HSE, a local search over encodings, and a check of the symmetry predictions.

**Reproducibility.** The code and data behind every table and figure are archived at https://doi.org/10.5281/zenodo.24061785: mhse_check.py, group_check.py, grid_n3.py, uniform_check.py, cj_table.py and cj_group.py for the $n = 3$ results, make_figures.py for the figures, larger_n.py for every result at $n = 4$ to 6 and for the numerical checks of Section 5 (sub-command theory), together with the stored outputs of each run. The archive includes a regression test (larger_n.py check) that recomputes the stored values.

## 7 RESULTS

### 7.1 Hamming sphere encodings are the easiest classes below p = 0.5

Table 2 reports, for the GA with fitness-proportional selection, the range of expected hitting times over the HSE classes and the best value achieved by any other landscape. In every setting with $p < 0.5$ and every crossover option, the fastest classes are all and only the HSE classes: the 6 (or 18) fastest classes are HSE, and the best landscape outside the class is slower than the slowest HSE. The margin is largest at low mutation rates (for example 12.004 versus 12.662 expected generations at $p = 0.05$ with one-point crossover) and shrinks as p approaches 0.5, where all landscapes approach random search.

Table 2: Expected generations to the optimum for the two-member GA with fitness-proportional selection, $n = 3$, $p_c = 0.8$ when crossover is used. "HSE" gives the range over all HSE classes; "best other" is the fastest landscape that is not an HSE. * marks settings ($p > 0.5$) where some non-HSE landscape is faster than every HSE; there the HSE are the slowest classes (Section 7.5).

| p | None: HSE | None: best other | One-point: HSE | One-point: best other | Uniform: HSE | Uniform: best other |
|---|---|---|---|---|---|---|
| 0.02 | 29.477 to 30.172 | 30.892 | 26.973 to 27.690 | 29.054 | 27.446 to 28.103 | 29.497 |
| 0.05 | 13.027 to 13.224 | 13.670 | 11.792 to 12.004 | 12.662 | 12.031 to 12.215 | 12.924 |
| 0.10 | 7.421 to 7.478 | 7.806 | 6.744 to 6.818 | 7.098 | 6.887 to 6.941 | 7.248 |
| 0.20 | 4.568 to 4.578 | 4.695 | 4.288 to 4.310 | 4.412 | 4.356 to 4.366 | 4.482 |
| 1/3 | 3.514 to 3.515 | 3.555 | 3.440 to 3.447 | 3.482 | 3.461 to 3.461 | 3.503 |
| 0.45 | 3.265 to 3.265 | 3.275 | 3.259 to 3.259 | 3.269 | 3.261 to 3.261 | 3.271 |
| 0.55 | 3.327 to 3.327 | 3.043 * | 3.321 to 3.321 | 3.039 * | 3.323 to 3.323 | 3.041 * |
| 0.65 | 3.651 to 3.651 | 2.804 * | 3.591 to 3.596 | 2.772 * | 3.607 to 3.608 | 2.786 * |

The same holds for the (1+1) EA (Table 3) and for tournament selection (Table 4). For the (1+1) EA the six HSE classes take the same time, as Proposition 2 requires, and the closest competitor is less than 0.2% slower at $p = 0.1$. With tournament selection the HSE again occupy all the top classes below 0.5.

Table 3: (1+1) EA, $n = 3$. All HSE share one value (Proposition 2). $\rho$ is the Spearman correlation with expected hitting time over all 40,320 one-to-one three-bit landscapes.

| p | HSE value | Best other | Position of HSE | ρ (violations) | ρ (FDC) |
|---|---|---|---|---|---|
| 0.02 | 60.234 | 60.237 | easiest 6 of 840 | 0.80 | 0.79 |
| 0.05 | 25.307 | 25.316 | easiest 6 of 840 | 0.80 | 0.79 |
| 0.10 | 13.775 | 13.795 | easiest 6 of 840 | 0.80 | 0.79 |
| 0.20 | 8.275 | 8.315 | easiest 6 of 840 | 0.81 | 0.80 |
| 1/3 | 6.597 | 6.637 | easiest 6 of 840 | 0.82 | 0.82 |
| 0.45 | 6.645 | 6.654 | easiest 6 of 840 | 0.83 | 0.83 |
| 0.55 | 7.610 | 5.898 | hardest 6 of 840 | -0.83 | -0.83 |
| 0.65 | 10.071 | 4.849 | hardest 6 of 840 | -0.81 | -0.82 |

Table 4: Two-member GA with binary tournament selection, n = 3, at three of the eight mutation rates; the remaining rates behave the same way (HSE easiest for $p < 0.5$, hardest for $p > 0.5$).

| p | Crossover | HSE range | Distinct HSE values | Best other | HSE are |
|---|---|---|---|---|---|
| 0.05 | none | 12.925 to 12.925 | 1 | 13.125 | easiest |
| 0.05 | one-point | 12.221 to 12.235 | 14 | 12.396 | easiest |
| 0.05 | uniform | 12.365 to 12.365 | 1 | 12.544 | easiest |
| 0.10 | none | 7.559 to 7.559 | 1 | 7.693 | easiest |
| 0.10 | one-point | 7.147 to 7.158 | 14 | 7.267 | easiest |
| 0.10 | uniform | 7.236 to 7.236 | 1 | 7.357 | easiest |
| 0.65 | none | 3.421 to 3.421 | 1 | 2.839 | hardest |
| 0.65 | one-point | 3.394 to 3.395 | 14 | 2.817 | hardest |
| 0.65 | uniform | 3.401 to 3.401 | 1 | 2.822 | hardest |

### 7.2 Structure inside the class is explained by symmetry

Lemma 1 predicts how many distinct hitting times can occur, and the predictions are met (Table 5). With mutation only, the GA with fitness-proportional selection takes 840 distinct values, one per $B_3$-orbit, and the HSE take 6. With one-point crossover it takes 2,520 values, one per $H_3$-orbit, and the HSE take 18; within a $B_3$-orbit the spread reaches 0.71 generations at $p = 0.1$ and 6.97 at $p = 0.02$, while within an $H_3$-orbit it is at rounding level (about $10^{-13}$). Uniform crossover commutes with all of $B_3$, so it should restore 6 HSE values, and it does.

Table 5: Orbit structure of expected hitting times for n = 3 (fitness-proportional selection, $p_c = 0.8$). Within-orbit spreads are at rounding level (below $10^{-10}$) for the group listed.

| Algorithm (p = 0.1 unless stated) | Group | Predicted classes | Observed distinct values | Distinct HSE values |
|---|---|---|---|---|
| (1+1) EA | $B_3$ (order 48) | 840 | 610 | 1 |
| GA, mutation only | $B_3$ (order 48) | 840 | 840 | 6 |
| GA, one-point crossover | $H_3$ (order 16) | 2,520 | 2,520 | 18 |
| GA, one-point crossover, p = 0.02 | $H_3$ (order 16) | 2,520 | 2,520 | 18 |
| GA, uniform crossover | $B_3$ (order 48) | 840 | 840 | 6 |

So the additional differences that one-point crossover introduces among HSE are not noise. They are a positional (linkage) effect, the only kind of difference one-point crossover can introduce. The differences among the 6 $B_3$-classes themselves have a different source, fitness-proportional selection, which depends on the numeric fitness values; under tournament selection they disappear. Two further coincidences go beyond symmetry. The (1+1) EA takes only 610 distinct values instead of 840, and all HSE merge into one, which is the lumping of Proposition 2. With tournament selection, HSE take a single value under mutation only and under uniform crossover, which are Propositions 3 and 4, and at most 14 values instead of 18 under one-point crossover.

Earlier work on this problem found that the order of HSE among themselves changes with the mutation rate (for example, different directional HSE were best at $p = 0.05$ and $p = 0.45$ for $n = 5$). Table 2 shows that such orderings rest on small differences (1% to 3% at $n = 3$) with two sources: fitness-proportional selection, which separates the B_(n)-classes, and one-point crossover, which splits each class further by bit order. The first source has a structure of its own, which we examine next.

**Alignment of adjacent spheres.** Under mutation alone the expected time is constant on $B_n$-orbits, so whatever separates the HSE must be a property of the landscape that every bit permutation and every XOR mask preserves. The relative position of the values on neighbouring spheres is such a property. For adjacent spheres $S_j$ and $S_{j+1}$ about the optimum, take every edge of the hypercube joining them and correlate, over these edges, the rank of f at the endpoint in $S_j$ (among the values on $S_j$) with the rank of f at the endpoint in $S_{j+1}$. Call the result the alignment $a_{j,j+1}$. It is positive when the best points of one sphere neighbour the best points of the next, negative when they neighbour the worst, and it is defined for j = 1, …, n − 2, since $S_0$ and $S_n$ are single points. For the directional HSE of Appendix A the alignments are extreme: two spheres filled in the same numeric direction have $a_{j,j+1}$ = +0.59 or +0.66 at n = 5, and two filled in opposite directions have the negative of that.

Table 6 gives, for the GA with fitness-proportional selection and mutation only at n = 5, the correlation of expected time with each alignment over 400 HSE drawn at random, together with the fit of a linear model in the three alignments. Two regularities appear. First, the sign depends only on the distance from the optimum: alignment of the two spheres nearest the optimum ($a_{1,2}$) shortens the time, and alignment of every farther pair lengthens it. Second, which pair matters depends on the mutation rate. At p = 0.02 the nearest pair accounts for almost everything (r = −0.81), and at p ≥ 0.2 the farthest pair does (r = +0.83 to +0.91). This is why the best directional HSE changes with the rate: at low rates the winners align the two spheres nearest the optimum and alternate elsewhere (a/d/a/a and its mirror d/a/d/d at p = 0.05), and the losers do the opposite (a/a/a/d and d/d/d/a at every rate below 0.5). The linear model explains 69% to 87% of the within-class variance; the remainder is finer structure, since mirror pairs such as a/d/a/a and d/a/d/d have identical alignments but differ by 0.003 generations. The same pattern holds at n = 4, it survives the addition of uniform crossover ($R^2$ = 0.75 and 0.86 at p = 0.05 and 0.2, n = 5), and it holds when both members are mutated before selection instead of after. Under tournament selection without crossover every alignment is irrelevant, because all HSE take the same time (Proposition 3).

The alignment effect is not captured by Walsh energy aggregated by order. Hypercube-equivalent encodings share the magnitudes of their Walsh coefficients (Rowe et al., 2004), and so, at n = 5, do the directional HSE a/a/a/d and d/a/a/a, whose energy by order (the sum of squared coefficients over each order) is identical although, relative to a/a/a/a, the first is slower and the second faster at every rate below 0.5. Over 300 random five-bit HSE at p = 0.05, with fitness-proportional selection and mutation only as in Table 6, a linear model in the energies of orders 1 to 4 explains 11% of the variance in expected time, against 68% for the three alignments on the same sample.

Table 6: Expected time T of 400 random five-bit HSE against adjacent-sphere alignment (GA, fitness-proportional selection, mutation only). r is the Pearson correlation, $R^2$ that of the linear model in all three alignments. The last column names the fastest of the 16 directional HSE by the numeric direction used on the spheres of weight 1 to 4 (a ascending, d descending).

| p | r(T, $a_{1,2}$) | r(T, $a_{2,3}$) | r(T, $a_{3,4}$) | $R^2$ | Best directional HSE |
|---|---|---|---|---|---|
| 0.02 | −0.81 | +0.09 | +0.08 | 0.69 | d/a/a/a |
| 0.05 | −0.55 | +0.44 | +0.35 | 0.69 | a/d/a/a |
| 0.10 | −0.29 | +0.49 | +0.62 | 0.80 | a/d/a/a |
| 0.20 | −0.13 | +0.33 | +0.83 | 0.87 | d/a/d/d |
| 1/3 | −0.07 | +0.15 | +0.90 | 0.86 | d/a/d/d |
| 0.45 | −0.05 | +0.02 | +0.91 | 0.84 | d/a/a/a |

The mechanism can be read off the fundamental matrix. Flipping the direction of one sphere in the directional HSE a/a/a/a changes exactly one alignment, so the resulting change in expected time can be split by the state (i, j) in which the two members lie in $S_i$ and $S_j$. Flipping $S_1$ (giving a/a/a/d, which anti-aligns $S_1$ and $S_2$) costs 0.64 generations at p = 0.02, but the cost is not paid in $S_1$: the extra time is spent with both members in $S_2$ (+0.34) or in $S_3$ (+0.15). What changes in $S_1$ is the risk of leaving it. With the strong points of $S_1$ now adjacent to the weak points of $S_2$, a member of $S_1$ faces a relatively stronger partner, and the probability per generation that the population drops out of $S_1$ rises from 0.0743 to 0.0753. Over a run this is 0.03 additional losses, each costing about 20 generations to regain $S_1$ at p = 0.02. Alignment near the optimum therefore acts by fixing forward steps. Flipping $S_4$ (giving d/a/a/a, which anti-aligns $S_3$ and $S_4$) saves time in the states (3, 4) and (2, 4) that follow a backward step: a member that has fallen from $S_3$ to $S_4$ is eliminated in the next generation with probability 0.635 instead of 0.597, at the price of slightly more time with both members in $S_4$, because forward steps out of $S_4$ now land on weak points and are fixed less often. Anti-alignment far from the optimum therefore acts by undoing backward steps. Both effects exist for every pair of spheres; near the optimum fixing a forward step is what matters, since the next step is absorption, and far from it undoing a backward step is. The rate dependence in Table 6 follows: the far effect is nearly constant in absolute terms (0.03 to 0.05 generations at $p = 0.02$ to 0.2), while the cost of losing $S_1$ falls from about 20 generations at $p = 0.02$ to about 5 at $p = 0.2$, and with it the near effect, from 0.64 to 0.013 generations. The frequency of backward steps is not the cause; a run contains 3.6, 3.3, 3.3 and 4.4 of them at $p = 0.02$, 0.05, 0.2 and 0.45.

### 7.3 The best and worst landscapes

Figure 1 shows the 16 fastest $B_3$-classes for the GA with uniform crossover at $p = 0.05$, drawn as cubes with the optimum (value 7) at the back top right. Uniform crossover was chosen because it commutes with every cube symmetry, so each drawing stands for a single class. Classes #1 to #6 are the six HSE classes of Table 1 ($v = 0$, where $v$ is the number of sphere-order violations defined in Section 7.4). Classes #7 to #16 all have a single violation: one value from the first sphere trades places with one from the second.

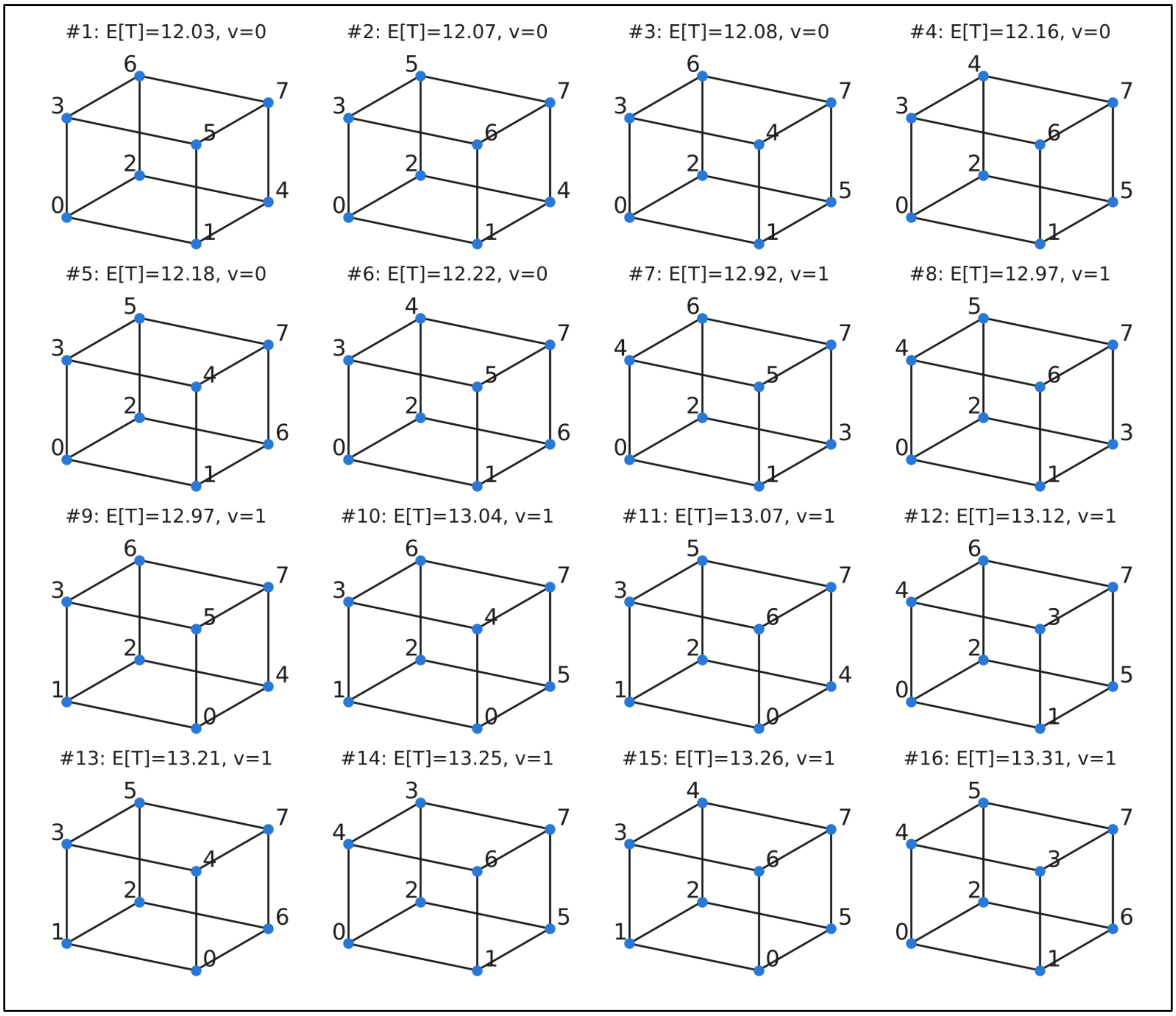

Figure 1. The 16 fastest classes of three-bit landscapes (two-member GA, fitness-proportional selection, uniform crossover $p_c = 0.8$, $p = 0.05$). Vertex labels are fitness values; 7 is the optimum. #k marks the k-th fastest class. E[T] is the expected number of generations; v is the number of sphere-order violations.

Figure 2 shows the 16 slowest classes, with v between 13 and 15 (15 is the maximum for n = 3). In all 16, the three neighbors of the optimum carry the three lowest values; in the six slowest, the second-best value (6) also sits at the antipode 000. These are trap functions: away from the optimum the landscape is an MHSL centred on the complement of x*. This is the deceptive structure that He et al. (2015) identify as hardest for the (1+1) EA, and it is a maximal violation of the sphere order rather than a slight one.

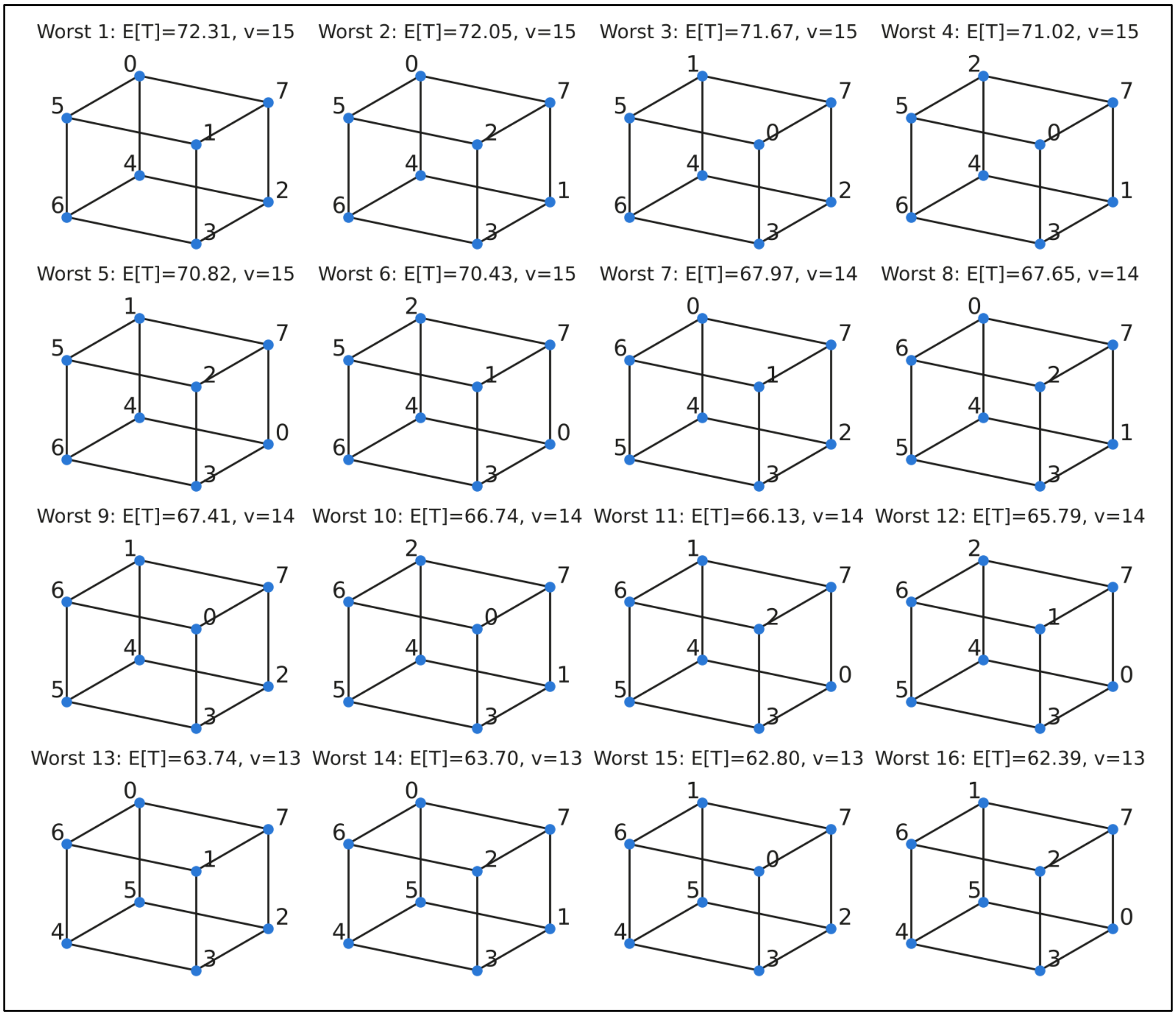

Figure 2. The 16 slowest classes for the same GA as Figure 1, from the slowest (Worst 1) upward.

### 7.4 Counting sphere-order violations

For a landscape f with optimum $x^*$, let

$$v(f) = |\{(x, y) : d(x) < d(y) \text{ and } f(x) < f(y)\}|,$$

the number of pairs whose order contradicts the sphere order. For the one-to-one landscapes studied here, $v = 0$ exactly when f is an HSE, so v measures how far a landscape is from the class; the trap of Figure 2 attains the maximum. The statistic is not new: v counts the discordant pairs between fitness and distance, so for fixed n Kendall's $\tau_b$ between f and d equals $(2v - P)/\sqrt{(NP)}$, where P is the number of pairs at different distances from $x^*$ and $N = 2^n(2^n - 1)/2$ is the number of all pairs. The two therefore rank landscapes identically. Like FDC, v requires the optimum, so it is a statistic for analyzing a landscape whose optimum is known or estimated, not a predictor for unseen problems; He et al. (2015) note that no efficient general predictor exists. Figure 3(a) plots expected hitting time against v for all 840 classes; hitting time rises steadily and almost monotonically with v. Table 7 gives the Spearman correlations: below p = 0.5 they lie between 0.92 and 0.99, and above 0.5 they change sign, consistent with Section 7.5. FDC, the Pearson correlation between f and d (Jones & Forrest, 1995), gives nearly the same values, as expected, since both statistics measure agreement between fitness and distance. The same holds at n = 5, where both correlations lie between 0.88 and 0.94 below p = 0.5 (Section 7.6).

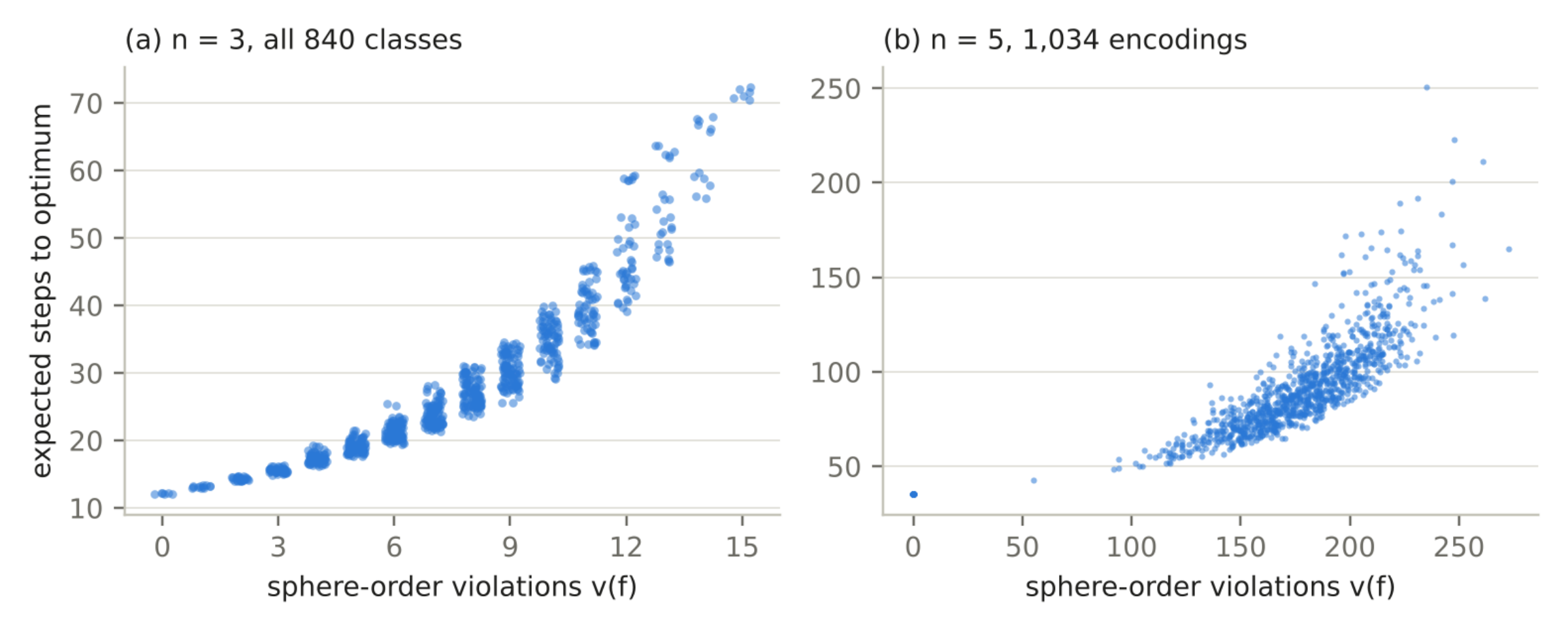

Figure 3. Expected steps to the optimum against the number of sphere-order violations v(f). (a) All 840 three-bit classes, same GA as Figure 1 (points jittered horizontally). (b) The 1,034 five-bit encodings of Table 8 at p = 0.05 (Section 7.6).

Table 7: Spearman correlation of expected hitting time with the violation count v / with FDC, over all 40,320 one-to-one three-bit landscapes (GA, fitness-proportional selection). Because FDC is negative for easy landscapes, both columns are oriented so that positive values mean the measure orders landscapes the same way as difficulty.

| p | None | One-point | Uniform |
|---|---|---|---|
| 0.02 | 0.97 / 0.95 | 0.97 / 0.96 | 0.97 / 0.96 |
| 0.05 | 0.98 / 0.97 | 0.98 / 0.98 | 0.98 / 0.98 |
| 0.10 | 0.99 / 0.98 | 0.98 / 0.98 | 0.98 / 0.98 |
| 0.20 | 0.97 / 0.96 | 0.96 / 0.96 | 0.96 / 0.96 |
| 1/3 | 0.94 / 0.94 | 0.93 / 0.93 | 0.93 / 0.93 |
| 0.45 | 0.92 / 0.92 | 0.92 / 0.92 | 0.92 / 0.92 |
| 0.55 | -0.91 / -0.92 | -0.91 / -0.92 | -0.91 / -0.92 |
| 0.65 | -0.91 / -0.93 | -0.90 / -0.92 | -0.90 / -0.92 |

### 7.5 Mutation rates above one half

At p = 0.55 and p = 0.65 the result reverses completely: in every setting tested (the (1+1) EA, both selection schemes and all three crossover options), the HSE classes are the *slowest* classes, and the correlations of Table 7 become negative. Section 5 gives the reason for the (1+1) EA. Above 0.5 a mutation flips most bits, so offspring tend to land near the complement of the parent, and points far from the optimum become the best starting points. The GA inherits this behaviour from its mutation operator. The claims of this paper are therefore claims about p < 0.5, which includes the rates ordinarily used in bit-string evolutionary algorithms. Rates above 0.5 are rarely studied, although Kelley (2026) shows that for every constant rate, including rates above 0.5, there is a function on which that rate is optimal for the (1+1) EA; the inversion reported here is the complementary observation that above 0.5 the landscapes that are easiest below 0.5 become the hardest.

It is also a natural instance of a result of He et al. (2015), who prove that the easiest function for one elitist (1+1) EA can be the hardest for another by constructing an artificial mutation operator; here the second algorithm is the same operator with p above 0.5. We include these rates only to show that the condition $p < 0.5$ in Corollary 1 is tight.

### 7.6 Larger n

Exhaustive evaluation is currently impractical beyond n = 3 ($16! \approx 2.1 \times 10^{13}$ landscapes, or about $5.4 \times 10^{10}$ classes, at n = 4). For n = 4 to 6 we therefore use four targeted tests with the GA of Section 6 (fitness-proportional selection, $p_c$ = 0.8, unless stated otherwise): a comparison with other encodings, perturbations of HSE, a local search over all encodings, and a check of the symmetry predictions of Section 4.

**Comparison with other encodings.** For n = 5 we evaluated 1,034 encodings at p = 0.05, 0.45 and 0.65 with uniform crossover: the 16 directional HSE of Appendix A, 16 HSE drawn at random (random optimum, random order inside each sphere), standard binary, the binary-reflected Gray code and 1,000 random permutations (Table 8). The outcome matches n = 3. At p = 0.05 and 0.45 the 32 HSE are the 32 fastest encodings, and at p = 0.65 they are the 32 slowest. Below 0.5 the best encoding outside the class is standard binary, which at p = 0.05 is 20% slower than the slowest HSE (42.66 against 35.55 generations). The violation count correlates with hitting time at ρ = 0.89 and 0.93 below 0.5, and at −0.82 at p = 0.65.

Table 8: Five-bit results for 1,034 encodings (GA, fitness-proportional selection, uniform crossover, $p_c$ = 0.8). Correlations are oriented as in Table 7.

| p | 32 HSE | Best non-HSE | HSE positions among 1,034 | ρ (v) |
|---|---|---|---|---|
| 0.05 | 35.28 to 35.55 | 42.66 (standard binary) | 1 to 32 | 0.89 |
| 0.45 | 14.568 to 14.569 | 14.77 (standard binary) | 1 to 32 | 0.93 |
| 0.65 | 18.47 to 18.48 | 13.82 (random permutation) | 1,003 to 1,034 | −0.82 |

**Perturbations of HSE.** To test how sharp the class boundary is, we perturbed each directional HSE by swapping the fitness values of two points on different spheres, never moving the optimum. For n = 5 (uniform crossover, p = 0.05), each of the 16 directional HSE has 355 such single swaps, and we also sampled 200 double swaps per HSE that do not restore sphere order. All 5,680 single swaps and 3,200 double swaps are slower than the HSE they were made from, so every directional HSE is a strict local optimum of hitting time under these moves. The margin is small, however. The fastest single swap of each HSE is only 0.024 to 0.053 generations slower than its base, whereas the directional HSE themselves span 0.28 generations (35.28 to 35.55). Some perturbed landscapes are therefore faster than some HSE: the fastest single swap overall (35.303) beats 14 of the 16 directional HSE. Unlike at n = 3, at n = 5 it is not true that every HSE beats every other landscape.

At n = 4 and p = 0.05 the boundary is still sharp with crossover. With one-point crossover, uniform crossover, or tournament selection (with uniform crossover), every single swap and every sampled double swap of every directional HSE is slower than all eight directional HSE. Without crossover, each swap is still slower than its own base, but some beat other HSE, as at n = 5.

**Local search.** We ran a hill climber over encodings from random permutations, using swap moves with first improvement, until no swap improved the expected time. At n = 5 (uniform crossover, p = 0.05) all 10 runs (2,094 to 4,078 evaluations each) ended at an HSE. The 10 end points were all different HSE, with times from 35.2705 to 35.2746, and each was faster than the best directional HSE (35.2772). At n = 4 and p = 0.05, all 10 runs ended at an HSE for each of five variants: mutation only, one-point crossover, uniform crossover, tournament selection, and the (1+1) EA. The same held at p = 0.2 with uniform crossover. At p = 0.65 no run ended at an HSE. All five runs reached the same time (6.29), on landscapes where the four neighbors of the optimum carry the four lowest values, which is the trap structure of Figure 2. Starting from arbitrary landscapes, the search therefore finds HSE at both rates below 0.5 that we tested. This supports, but does not prove, the claim that the fastest landscapes at n = 4 and 5 are HSE, and it shows that the directional HSE of Appendix A are not the best HSE.

**Symmetry.** For three random HSE at n = 4 we evaluated all 384 images of each under $B_4$. With no crossover or with uniform crossover all images have the same expected time; with one-point crossover there are exactly 12 distinct values, which is $|B_4|/|H_4| = 384/32$, the most that Lemma 1 allows. The same counts hold with tournament selection.

**Six bits.** For the 32 directional HSE at n = 6 ($p = 0.02$, $p_c = 0.8$), expected times span 132.38 to 133.22 generations with one-point crossover and 131.08 to 131.95 with uniform crossover. These are spreads of 0.6% and 0.7%, consistent with the small within-class differences at n = 3.

## 8 DISCUSSION

**Why spheres.** For mutation-based search, what matters about a point is its distance to the optimum: bitwise mutation moves between spheres with probabilities that depend only on the distances involved. A landscape that makes every closer sphere fitter than every farther one never rejects a move that gets closer and never accepts one that moves away, so selection always pushes in the right direction. Proposition 2 establishes this equivalence for the (1+1) EA, and Proposition 3 and Theorem 1 extend it, with the optimality, to the two-member GA with tournament selection and no crossover. With fitness-proportional selection or crossover the equivalence does not follow, because selection then depends on the numeric values and crossover on more than distances. The three-bit results nevertheless show that it survives as an ordering: sphere order decides which landscapes are easiest, and the remaining details only reorder them inside the class. The contrast between the two selection schemes separates ordinal from cardinal structure: tournament selection sees only the order the landscape induces, which enables Proposition 3 and, together with the injectivity of HSE and the symmetry of uniform crossover, Proposition 4, whereas fitness-proportional selection also sees the spacing of the values, which is what separates the HSE in Section 7.2.

**What crossover adds.** Crossover contributes in two ways: it changes which symmetries the algorithm respects, and it can combine correct bits carried by different members (Section 5). Uniform crossover respects every symmetry of the cube and does not separate HSE that mutation treats alike. One-point crossover depends on bit adjacency and separates them by small amounts. At n = 3 these effects are an order of magnitude smaller than the gap to non-HSE landscapes. At n = 5 they are not: the directional HSE differ by more than the gap between an HSE and its nearest perturbations (Section 7.6).

**Neutrality.** The free order inside spheres does not make the (1+1) EA slower. Ties *across* spheres can, because they allow equally fit moves away from the optimum. This refines the original motivation for MHSL as a model of tiered problems: the tiers themselves are harmless; plateaus that join tiers are what costs time. In the tested two-member tournament GA, ties inside a tier can even help when crossover is present: OneMax was faster than every HSE in the reported settings (Section 5).

**Hardest landscapes.** The slowest three-bit landscapes are traps, in agreement with He et al. (2015). They are maximal, not slight, violations of sphere order.

**Limitations.** Exhaustive results are for n = 3 and a two-member population. Fitness-proportional selection makes absolute times depend on the numeric fitness values, although the tournament results show that the main conclusion does not. The two-member GA is non-elitist, and its hitting time grows rapidly with n (Section 5); an elitist GA or a larger population would behave differently. Evidence for larger n is targeted rather than exhaustive. At n = 5 every directional

HSE is a local optimum under value swaps and every local-search run ends at an HSE, but some perturbed landscapes are faster than some HSE, so the three-bit statement that every HSE beats every other landscape does not carry over unchanged.

**Extensions.** Although our definition of MHSL requires a unique optimum, relaxing this constraint to allow flat peaks would admit further threshold and plateaued unitation functions; characterizing the resulting isotropic class is a natural extension. Other open questions are proofs of Conjectures 1 and 2, the latter extending Theorem 1 to crossover and to fitness-proportional selection, experiments with larger populations and with two-point crossover, which respects a different subgroup of $B_n$, and a landscape measure that captures the within-class differences of Section 7.2 better than the Walsh spectrum does. Dynamic representation methods such as the Gray-code shifting of Rowe et al. (2004) also move the induced landscape through different assignments of phenotype values to bit strings; whether the resulting change in sphere-order violations predicts when a shift helps is worth testing.

## 9 CONCLUSION

Monotonic Hamming sphere landscapes, the binary landscapes ordered by distance to a unique optimum, form a small class that contains OneMax and whose size we determine. For the elitist (1+1) EA, every sphere-separated MHSL has the same optimization-time distribution as OneMax, and known theory then makes these landscapes the easiest for mutation rates below 0.5. For a two-member genetic algorithm with tournament selection and no crossover we prove the same: every sphere-separated MHSL has OneMax's distance chain and is an easiest landscape below 0.5. For the remaining settings, exhaustive Markov-chain analysis of all three-bit landscapes shows the same picture: the easiest classes below 0.5 are all and only the Hamming sphere encodings, and above 0.5 the hardest classes are. Differences inside the class are small and come from the selection scheme and from which symmetries of the cube the operators respect. The number of sphere-order violations, which measures how far a landscape is from the class, correlates closely with hitting time on three and five bits. Together, these results identify sphere order, rather than standard binary or Gray coding, as the property that makes binary landscapes easy for evolutionary search at ordinary mutation rates.

## APPENDIX A: ENUMERATING DIRECTIONAL HAMMING SPHERE ENCODINGS

The procedure below generates HSE by visiting the strings of each Hamming weight in increasing or decreasing numeric order (the *direction* chosen for each sphere) and assigning consecutive values. With $2^{n-1}$ direction vectors it produces the directional HSE used in Sections 7.2 and 7.6 (4 for n = 3, 32 for n = 6); the function randomize_layers in the accompanying code permutes values inside spheres to sample other HSE. Here the optimum is $1^n$ and the fitness value of x is positions[x].

**Notation.** bitsize is the number of bits; k is the number of ones; positions[] holds the fitness value of each string; directions[] gives the enumeration direction for each weight; next_number(n) and previous_number(n) return the next larger and next smaller integer with the same number of ones; n_choose_k(n, k) is the binomial coefficient.

```
procedure next_number(n)
---
  c ← n AND (−n)                ▻ rightmost set bit of n
  r ← n + c                     ▻ flip rightmost non-trailing zero
  return (((r XOR n) >> 2) / c) OR r
end procedure
procedure previous_number(n)
  y ← (NOT n) AND (−(NOT n))    ▻ lowest zero bit of n
  n ← n AND NOT (y − 1)         ▻ reset bits below y
  z ← n AND (−n)                ▻ lowest set bit after reset
  n ← n AND NOT z               ▻ clear bit z
  n ← n OR (z − z / (2 × y))    ▻ refill bits below z
  return n
end procedure
procedure enumerate_by_hamming_weight(bitsize, directions[])
  positions[0 … 2^bitsize − 1] ← 0
  counter ← 1
  for k from 1 to bitsize do
    count ← n_choose_k(bitsize, k)
    value ← (1 << k) − 1
    dir ← directions[min(k − 1, bitsize − 2)]
    if dir = descend then value ← value << (bitsize − k)
    for step from 1 to count do
      positions[value] ← counter;  counter ← counter + 1
      if dir = descend then value ← previous_number(value)
      else value ← next_number(value)
      if value ≥ 2^bitsize or value = 0 then exit loop
    end for
  end for
  return positions
end procedure
procedure main()
```

```
    for each directions[] in {ascend, descend}^(bitsize − 1) do
        print enumerate_by_hamming_weight(bitsize, directions)
end procedure
```

## APPENDIX B: GROWTH OF THE NUMBER OF HAMMING SPHERE ENCODINGS

Taking logarithms in Proposition 1 and applying Stirling's approximation log m! ≈ m log m − m to each factor gives

$$\log E_n \approx n \log 2 + \Sigma_k C(n,k) \log C(n,k) - 2^n.$$

Writing $C(n,k) = 2^n \pi_k$, where $\pi_k$ is the Binomial(n, 1/2) probability with entropy $h_n = -\Sigma\, \pi_k \log \pi_k \approx ½ \log(\pi en/2)$, the middle term equals $2^n(n \log 2 - h_n)$. Hence

$$\log E_n \approx 2^n (n \log 2 - ½ \log(\pi en/2) - 1) + n \log 2,$$

so $E_n$ grows like $\exp(2^n\, n \log 2\, (1 + o(1)))$, which is doubly exponential in n. The number of all landscapes satisfies $\log (2^n)! \approx 2^n(n \log 2 - 1)$, so the fraction of landscapes that are HSE decays roughly like $\exp(-2^{n-1} \log(\pi en/2))$. Values are given in Table 9.

Table 9: Counts of Hamming sphere encodings.

| n | HSE (E_n) | Classes up to B_n | All landscapes (2^n)! | Fraction |
|---|---|---|---|---|
| 2 | 8 | 1 | 24 | 0.33 |
| 3 | 288 | 6 | 40,320 | $7.1 \times 10^{-3}$ |
| 4 | 6,635,520 | 17,280 | $2.09 \times 10^{13}$ | $3.2 \times 10^{-7}$ |
| 5 | $6.07 \times 10^{18}$ | $1.58 \times 10^{15}$ | $2.63 \times 10^{35}$ | $2.3 \times 10^{-17}$ |
| 6 | $1.38 \times 10^{50}$ | $3.00 \times 10^{45}$ | $1.27 \times 10^{89}$ | $1.1 \times 10^{-39}$ |

## APPENDIX C: CLASSES OF THREE-BIT FUNCTIONS BY LOCAL MINIMA

Table 10 lists, for all 40,320 functions on three bits, the number of functions with a given number of local minima in the integer (ring), Gray and binary neighborhoods, recomputed with the script cj_table.py. The grouping is the one used by Chakraborty and Janikow (2003b, Table 4); the sizes agree with theirs. The last column shows that every class size is a multiple of 16, the order of the group described in Section 4.3.

Table 10: Number of three-bit functions by number of local minima under the three neighborhoods. Each size is 16 × an integer.

| Integer | Gray | Binary | Functions | Functions / 16 |
|---|---|---|---|---|
| 1 | 1 | 1 | 64 | 4 |
| 1 | 1 | 2 | 384 | 24 |
| 1 | 1 | 3 | 64 | 4 |
| 2 | 1 | 1 | 768 | 48 |
| 2 | 1 | 2 | 5,248 | 328 |
| 2 | 1 | 3 | 128 | 8 |
| 2 | 2 | 1 | 2,288 | 143 |
| 2 | 2 | 2 | 4,784 | 299 |
| 2 | 2 | 3 | 1,232 | 77 |
| 2 | 2 | 4 | 144 | 9 |
| 3 | 1 | 1 | 224 | 14 |
| 3 | 1 | 2 | 1,568 | 98 |
| 3 | 1 | 3 | 192 | 12 |
| 3 | 2 | 1 | 2,016 | 126 |
| 3 | 2 | 2 | 9,024 | 564 |
| 3 | 2 | 3 | 3,664 | 229 |
| 3 | 2 | 4 | 1,296 | 81 |
| 3 | 3 | 1 | 1,872 | 117 |
| 3 | 3 | 2 | 2,704 | 169 |
| 3 | 3 | 3 | 480 | 30 |
| 4 | 2 | 1 | 16 | 1 |
| 4 | 2 | 2 | 16 | 1 |
| 4 | 3 | 1 | 432 | 27 |
| 4 | 3 | 2 | 272 | 17 |
| 4 | 4 | 1 | 960 | 60 |
| 4 | 4 | 2 | 480 | 30 |